\documentclass[letterpaper, 10 pt, conference]{ieeeconf}

\IEEEoverridecommandlockouts                            
\usepackage{graphics} 
\usepackage{amsmath} 
\usepackage{amssymb} 
\usepackage{balance}
\usepackage{mathtools}
\usepackage{xcolor}
\usepackage{graphicx}
\usepackage[font=footnotesize]{caption}
\def\transition{\Delta}
\def\horizon{T}
\def\weights{w}

\newcommand{\argmax}{\operatornamewithlimits{argmax}}

\DeclarePairedDelimiter\norm{\lVert}{\rVert}

\usepackage{xcolor}
\definecolor{customcolor}{RGB}{153,27,30}
\usepackage[hidelinks]{hyperref}
\hypersetup{
    colorlinks=true,
    linkcolor=customcolor,
    filecolor=magenta,      
    urlcolor=customcolor,
    citecolor=customcolor,
}

\usepackage{titlesec}
\titlespacing*{\subsection}{0pt}{0.3\baselineskip}{0.1\baselineskip}
\titlespacing*{\section}{0pt}{0.6\baselineskip}{0.5\baselineskip}

\usepackage[style=ieee,mincitenames=1,maxcitenames=2,maxbibnames=20]{biblatex}
\let\citet\textcite
\title{\LARGE \bf
A Generalized Acquisition Function\\for Preference-based Reward Learning
}

\author{Evan Ellis$^{1}$, Gaurav R. Ghosal$^{1,2}$, Stuart J. Russell$^{1}$, Anca Dragan$^{1}$, Erdem B\i y\i k$^{1,3}$
\thanks{This work was supported by Open Philantrophy and ONR Young Investigator Program (YIP).}
\thanks{Email addresses: \{\texttt{evan.ellis}, \texttt{gauravrghosal}, \texttt{russell}, \texttt{anca}\}\texttt{@berkeley.edu}, \texttt{biyik@usc.edu}}
\thanks{$^{1}$Department of Electrical Engineering and Computer Sciences, UC Berkeley}%
\thanks{$^{2}$Machine Learning Department, Carnegie Mellon University}%
\thanks{$^{3}$Thomas Lord Department of Computer Science, University of Southern California}%
}

\newtheorem{remark}{Remark}

\begin{document}

\maketitle
\thispagestyle{empty}
\pagestyle{empty}

\begin{abstract}
    Preference-based reward learning is a popular technique for teaching robots and autonomous systems how a human user wants them to perform a task. Previous works have shown that actively synthesizing preference queries to maximize information gain about the reward function parameters improves data efficiency. The information gain criterion focuses on precisely identifying all parameters of the reward function. This can potentially be wasteful as  many parameters may result in the same reward, and many rewards may result in the same behavior in the downstream tasks. Instead, we show that it is possible to optimize for learning the reward function up to a behavioral equivalence class, such as inducing the same ranking over behaviors, distribution over choices, or other related definitions of what makes two rewards similar. We introduce a tractable framework that can capture such definitions of similarity. Our experiments in a synthetic environment, an assistive robotics environment with domain transfer, and a natural language processing problem with real datasets demonstrate the superior performance of our querying method over the state-of-the-art information gain method.
\end{abstract}

\section{Introduction}
Reliably inducing desirable behavior in robots is an important prerequisite for their real-world deployment. As desirability is fundamentally human-dependent and subjective, prior work has extensively studied the problem of learning reward functions from human feedback. Although this approach is intuitive, it can be costly to collect sufficient human data to identify a faithful reward function. Moreover, errors in the identified reward function can often lead to highly undesirable consequences \cite{skalse2022defining}.

A popular approach for improving data efficiency in reward learning is active learning, where a robot generates the queries that are the most informative for identifying the reward function. This approach focuses human labeling efforts on resolving presently under-specified aspects of the reward function, making it appealing for avoiding hard-to-detect misalignment in reward functions. Prior works have proposed various objectives for quantifying how informative a query is, such as the volume removed from the robot's belief over the reward function \cite{sadigh2017active} or the information gain about the reward function \cite{biyik2019asking}. 

Existing objectives such as volume removal and information gain optimize for reducing the uncertainty over the reward function parameters. However, what we care about is not the exact weights over features, but instead the correctness of the reward function for the downstream task, as measured by the behavior it induces -- be this the distribution over trajectories, how possible candidate trajectories rank or compare, or what the optimal trajectory/policy is. 
These characteristics are rarely unique in parameter space, and there are often families of reward parameters that are indistinguishable in terms of these properties \cite{ng1999policy,skalse2023invariance}. Previous methods ignore this fact and often ask queries that yield little benefit in the downstream application.

Our key insight in this work is that \emph{the active learning algorithm should encourage learning the true reward function only up to an equivalence class of statistics over the induced behavior.
} Doing so enables the algorithm to focus on learning what matters, such as how the true reward ranks trajectories, and skip questions that are irrelevant to this objective.

To this end, we introduce a novel framework that allows active learning policies to focus on learning the true reward function for an \emph{alignment} metric that captures the functional characteristics we care about when comparing rewards. This alignment metric can leverage prior works in reward distance metrics \cite{jenner2022a} and can encode distributional information about the domain where we wish to deploy our reward function. Despite the flexibility of our framework, we provide a tractable approximation that holds under mild assumptions. 

To validate our approach, we run experiments on three different tasks: a synthetic environment, a simulated assistive robotics environment, and a natural language setting. We show results using three different measures of reward alignment, which induce different equivalent classes over the induced behavior: log-likelihood (does the reward choose the same query answer with the same probability?), EPIC distance \cite{gleave2021quantifying} (a state-of-the art metric for alignment that goes beyond induced optimal policies), and trajectory rankings (does the reward rank trajectories in the same order with the same magnitudes?). We outperform state-of-the-art performance by up to 85\% in learning rewards that transfer well to new domains, using both linear and nonlinear rewards.

\section{Related Work}
\textbf{Reward Learning from Human Feedback.} Hand-designing a reward function that induces desired behavior is generally challenging and prior works have consequently proposed methods for \emph{learning} reward functions from various forms of human input, such as demonstrations \cite{abbeel2004apprenticeship}, comparisons or rankings \cite{sadigh2017active,christiano2017deep,myers2021learning}, physical corrections \cite{bajcsy2017learning, bajcsy2018learning}, and emergency stops \cite{hadfield2017off}. In this work, we primarily consider learning from pairwise trajectory comparisons, although our work generalizes to other feedback types when they can be modeled appropriately. 

\noindent\textbf{Active Reward Learning.} Data efficiency is crucial in reward learning, as human feedback is typically costly to obtain. \emph{Active learning} techniques seek to alleviate this burden by identifying the queries with the most informative feedback. \cite{mackay1992information}. Prior works have considered multiple metrics for query informativeness including the volume removed from the hypothesis space \cite{sadigh2017active} and mutual information \cite{biyik2019asking}. Recently, \citet{wilde2020active} demonstrated existing reward learning objectives that largely depend on parameter-space uncertainty can be suboptimal and proposed a regret-based technique. However, this technique relies on the existence of an efficient method that provides the optimal policy for any given reward function. Similarly, \cite{lindner2022information} proposes to maximize information gain about the differences among a set of plausibly optimal policies. In this work, we extend their observations and provide a tractable and general method for addressing them.

\noindent\textbf{Reward Metrics.} While learned reward functions can be evaluated by computing the ground-truth returns of policies optimized on the learned reward \cite{christiano2017deep,wang2024rlvlmf}, this can introduce confounding factors from policy learning failures. An alternative approach is measuring the distance between the learned and ground-truth reward functions. Simple parameter-space metrics are typically insufficient as they correlate poorly with functional differences between reward functions \cite{jenner2022a}. For example, transformations of reward functions often yield the same optimal policy and ranking over trajectories \cite{ng1999policy,skalse2023invariance}. To address this, \citet{gleave2021quantifying} introduced the EPIC reward psuedometric which provably respects the equivalence class of reward functions inducing the same optimal policy. \citet{wulfe2022dynamicsaware} refined EPIC to better take into account the likelihood of transitions under an approximate dynamics model. Finally, \citet{balakrishnan2020efficient} introduced a reward function projection under which nearby points induce a similar likelihood on expert demonstrations. In this work, we demonstrate how reward metrics can be harnessed for efficient active reward learning.

\section{Problem Formulation}
We consider a Markov decision process represented with the tuple: $\langle \mathcal{S}, \mathcal{A}, \mu, \transition, r, \horizon \rangle$ where $\mathcal{S}$ and $\mathcal{A}$ are the state and the action spaces, respectively. The initial state of the system is distributed according to $\mu$, i.e., $s_0\sim\mu(\cdot)$. The distribution $\transition$ is the transition dynamics of the system such that when action $a_t\in\mathcal{A}$ is taken at state $s_t\in\mathcal{S}$ at timestep $t$, the next state is $s_{t+1} \sim \transition(\cdot \mid s_t, a_t)$. $\horizon$ represents the finite horizon. We denote the set of dynamically feasible trajectories as $\Xi$.

The reward function $r:\mathcal{S}\times\mathcal{A}\to\mathbb{R}$ is only known by a human user. We use $R$ to denote the cumulative reward over trajectories: $R(\xi) = \sum_{(s,a)\in\xi} r(s,a)$. The human prefers trajectories with higher cumulative rewards, i.e.,
\begin{align}
    \label{eq:preference_definition}
    \xi \succ \xi' \iff R(\xi) > R(\xi')
\end{align}
for any $\xi,\xi'\in\Xi$. Our goal is to efficiently learn the cumulative reward function $R$ from human feedback.

\noindent\textbf{Human Feedback.} We adopt pairwise comparison feedback where the human is presented with a pair of trajectories and is asked to select which one they prefer. We use $Q\in\Xi^2$ to denote the query and $q\in Q$ for the human's response to query $Q$. We assume access to a human response model conditioned on reward function, i.e., $P(q \mid Q, R)$. While we use this type of feedback in our experiments, our discussion of the existing solutions and the method we propose extend to any type of human feedback for which we have a human response model \cite{jeon2020reward}, such as rankings \cite{myers2021learning,brown2019extrapolating}, ordinal feedback \cite{chu2005gaussian,li2021roial}, and corrections \cite{bajcsy2017learning,bajcsy2018learning}.

\noindent\textbf{Learning from Human Feedback.} Given a dataset of query responses $\mathcal{D}_K\!=\!\{(Q_k,q_k)\}_{k=1}^{K}$, we learn a reward function $R_\weights\!:\!\Xi\!\to\!\mathbb{R}$, parameterized by $\weights$, in a Bayesian way:
\begin{align}
    p(\weights \!\mid\! \mathcal{D}_K) \!\propto\! p(\weights)p(\mathcal{D}_K \!\mid\! \weights) \!=\! p(\weights)\prod_{k=1}^K p(q_k \!\mid\! Q_k, R_\weights)
\end{align}
by assuming the human's responses are conditionally independent from each other given the reward function. While we use an uninformative prior $p(\weights)$ in our experiments, it is possible to inject domain knowledge or other forms of human data (e.g., demonstrations \cite{biyik2022learning}) into learning via the prior. Samples from the posterior distribution give estimates of the true reward function $R$.

\noindent\textbf{Adaptive Querying.} We are interested in an adaptive, human-in-the-loop setting, where we decide the query we will make to the human based on the previous queries and the human's answers to them:
\begin{align}
    Q_k = \pi(\mathcal{D}_{k-1})\:,
\end{align}
where $\pi$ is the adaptive querying policy.

\noindent\textbf{Objective.} Our goal is to actively learn a reward function $R_\weights:\Xi\to\mathbb{R}$ with a small number of queries such that the learned reward function aligns with the true reward function $R$ under some metric $f$. Mathematically, we want to find an adaptive querying policy by solving
\begin{align}
    \label{eq:objective}
     \mathrm{maximize}_\pi &\; \mathbb{E}_{\weights\sim P(\weights \mid \mathcal{D}_K)} f(R_\weights, R)\\
     \mathrm{s.t.} &\; Q_k = \pi(\mathcal{D}_{k-1})\; \textrm{for } k=1,2,\dots,K\nonumber
\end{align}

We start by reviewing the state-of-the-art active learning method in which an information-theoretic objective is greedily maximized. We then show its drawbacks, and propose our solution in Section~\ref{sec:approach}.

\section{Existing Solutions}
The state-of-the-art method for actively choosing queries is mutual information maximization \cite{biyik2019asking}. Given any dataset $\mathcal{D}_{k-1}$ of $k-1$ query-response pairs, the next query is selected to greedily maximize the mutual information between its expected response and the reward function parameters:
\begin{align}
    \label{eq:mutual_information_objective}
    Q^{\textrm{MI}}_k = \pi^{\textrm{MI}}(\mathcal{D}_{k-1}) = \argmax_{Q\in\Xi^2} I(q ; w \mid Q, \mathcal{D}_{k-1})\:.
\end{align}
The rationale behind this method is that the mutual information is the difference between prior and posterior entropies over $\weights$, and maximizing it is equivalent to minimizing the posterior entropy. Therefore, a query that has the highest mutual information is expected to decrease the uncertainty about the reward function parameters $\weights$ the most.

In fact, this approach outperformed the previous methods (e.g., volume removal \cite{sadigh2017active}) in various metrics, such as the cosine similarity between $\weights$ and $\weights^*$ \cite{sadigh2017active}, or the loglikelihood over a validation set of queries \cite{biyik2020active,biyik2023active}. However, \citet{wilde2020active} showed entropy minimization may not be the correct objective for the querying policy if the goal is not minimizing the entropy $H(\weights \mid \mathcal{D}_K)$. We describe their method and why $\pi^{\textrm{MI}}$ fails in the next section.

\section{Failure Cases of Mutual Information Based Querying Policy}
The interesting observation made by \citet{wilde2020active} is that when the goal is to find the optimal trajectory $\xi^*=\argmax_{\xi\in\Xi} R(\xi)$, the following maximum-regret based greedy optimization performs better than $\pi^{\textrm{MI}}$:
\begin{align}
    \label{eq:maximum_regret_policy}
    Q^{\textrm{MR}}_k &= \pi^{\textrm{MR}}(\mathcal{D}_{k-1}) = (\xi^{A},\xi^{B})\;\textrm{such that} \nonumber\\
    \xi^{A} &= \argmax_{\xi\in\Xi} R_{\weights^A}(\xi)\:,
    \xi^{B} = \argmax_{\xi\in\Xi} R_{\weights^B}(\xi)\:,\\
    \weights^A,\weights^B &= \argmax_{\weights^a,\weights^b} P(\weights^a \mid \mathcal{D}_{k-1})P(\weights^b \mid \mathcal{D}_{k-1})\nonumber\\
    &(R_{\weights^a}(\xi^A)-R_{\weights^a}(\xi^B) + R_{\weights^b}(\xi^B)-R_{\weights^b}(\xi^A))\:.\nonumber
\end{align}
Here, the regret is defined between two reward functions: if the true reward is parameterized by $\weights^A$ but the system learns $\weights^B$, then the regret is $R_{\weights^A}(\xi^A)-R_{\weights^A}(\xi^B)$ where $\xi^A$ and $\xi^B$ are the optimal trajectories under $R_{\weights^A}$ and $R_{\weights^B}$, respectively. This method implicitly makes the user choose between a pair of rewards, $R_{\weights^A}$ and $R_{\weights^B}$, which maximizes some regret metric by querying them with $(\xi^A,\xi^B)$.

The reason why maximum regret based policy $\pi^{\textrm{MR}}$ finds optimal trajectories better/faster than the mutual information based policy $\pi^{\textrm{MI}}$ is that the mutual information based method tries to greedily minimize the entropy over the reward function parameters, i.e., $H(\weights \mid \mathcal{D}_k)$. However, many different parameters may lead to the same optimal trajectory (reward ambiguity problem \cite{ng2000algorithms}). Therefore, the mutual information based policy $\pi^{\textrm{MI}}$ may make queries that are wasteful if the true goal is not to reduce entropy over parameters $\weights$. On the other hand, even if the goal is to find the optimal trajectory, the maximum regret based policy $\pi^{\textrm{MR}}$ has practical problems as the first two constraints in Equation~\eqref{eq:maximum_regret_policy} require the ability to optimize trajectories for any given reward function. These indicate a need for an efficient and general method that can handle various learning objectives, not just entropy minimization or trajectory optimization.

As another example of a common failure case of $\pi^{\textrm{MI}}$, suppose our goal in reward learning is to be able to compare any two trajectories in terms of their rewards, which is indeed true to the motivation of why reward functions exist. In such a case, if some parameters of the reward function are relevant only for a small subset of trajectories, then those parameters are less important than the others, as they will not affect most of the trajectory comparisons. However, $\pi^{\textrm{MI}}$ will give the same importance to all parameters, because they contribute to the entropy in the same way.

The mutual information based policy $\pi^{\textrm{MI}}$ is also not suitable for problems that involve domain transfer, i.e., the reward is learned in one domain and then transferred to the other. One may want to learn reward functions in simulation and then deploy the learned reward on a real robot due to safety concerns. Or it may be cheaper to collect human feedback in one domain than the other, e.g., if we think of each text as a trajectory a natural language processing (NLP) system takes, it is easier for most people to compare the writing quality of two paragraphs rather than two scientific articles. Due to the potential distribution shift between the domains, $\pi^{\textrm{MI}}$ will produce suboptimal queries as it is agnostic to the trajectory distribution of the systems.

These motivate us to develop our novel querying approach that lets designers plug their true alignment objective $f$.

\section{Our Approach}\label{sec:approach}
As we stated in Equation~\eqref{eq:objective}, our objective is to find an adaptive querying policy that maximizes $$\mathbb{E}_{\weights\sim P(\weights \mid \mathcal{D}_K)}\left[f(R_\weights,R)\right]$$
for some $f$ that captures the alignment between the true reward and the learned reward. However, in the most general case, we cannot compute this because we simply do not know the true reward function $R$. 

\noindent\textbf{Approximation. }Our approach is based on the observation that $P(\weights \mid \mathcal{D}_k)$ will give a high probability for $\weights^*$ that best aligns with the true reward, i.e.,
\begin{align}
    \weights^* = \argmax_\weights f(R_\weights,R) \:.
\end{align}
This observation holds under the mild assumption that $P(q \mid Q, R_{\weights^*}) \approx P(q \mid Q, R)$ for any query-response pair $(Q,q)$. Practically, this assumption only enforces that $f$ is really an alignment metric to be maximized, instead of adversarial metrics, e.g., one that is maximized when $R$ and $R_\weights$ make opposite predictions about user responses.

As a result of this observation, our insight is to solve
\begin{align}
    \label{eq:proxy_objective}
    \textrm{maximize}_\pi &\; \mathbb{E}_{\weights'\sim P(\weights\mid \mathcal{D}_K)}\left[\mathbb{E}_{\weights\sim P(\weights\mid \mathcal{D}_K)}\left[f(R_\weights,R_{\weights'})\right]\right]\\
    \mathrm{s.t.} &\; Q_k = \pi(\mathcal{D}_{k-1})\; \textrm{for } k=1,2,\dots,K\nonumber
\end{align}
as a proxy to the original problem, because the outer expectation in the objective is an expectation over cases where $\weights'$ is the target parameters $\weights^*$.

Intuitively, this gives birth to the notion of identifying the reward function up to a certain equivalence class. Say $f$ cared about the induced ranking over trajectories -- the original objective incentivizes that we find a reward which ranks trajectories similarly to the true reward; this proxy objective, which is computable based on the information we know, incentivizes that we identify the reward up to rewards that produce the same ranking; at that point, $f(R_w,R_{w'})$ becomes $0$ and further queries are no longer helpful.

\noindent\textbf{Greedy soluton. }Solving for the optimal adaptive policy $\pi^*$ is intractable, as it requires planning over $K$ queries each of which is answered stochastically by the user. We follow the prior work by taking a greedy approach to the problem. Merging the expectations in Equation~\eqref{eq:proxy_objective}, we let
\begin{align}
    \label{eq:greedy_objective}
    \pi^f(\mathcal{D}_{k-1}) =& \argmax_Q \mathbb{E}_{q\sim P(q \mid Q,\mathcal{D}_{k-1})}\big[\nonumber\\
    &\quad \mathbb{E}_{\weights,\weights'\sim P(\weights\mid \mathcal{D}_{k-1},Q,q)}
    \left[f(R_\weights,R_{\weights'})\right]\big]\:,
\end{align}
where we greedily optimize for the expected alignment of the posterior $P(\weights\mid \mathcal{D}_{k-1},Q,q)$ under $f$. The inner expectation is not trivial to compute, as it requires sampling parameter pairs from the posterior for the given query-response pair $(Q,q)$. Given the optimization is over $Q$, we would need to repeat sampling many times. However, as we show in the appendix, this greedy solution is simplified as:
\begin{align}
    \label{eq:final_objective}
    \argmax_Q \sum_{q\in Q}& \frac{\mathbb{E}_{\weights,\weights'}\left[P(q \mid Q, R_{\weights})P(q \mid Q, R_{\weights'})f(R_\weights,R_{\weights'})\right]}{\mathbb{E}_{\weights''} P(q \mid Q, R_{\weights''})}
\end{align}
where all expectations are over the prior, i.e., $P(\weights \mid \mathcal{D}_{k-1})$, and we only need to evaluate $f(R_\weights,R_{\weights'})$ and $P(q \mid Q, R_{\weights})$, which are already given. Therefore, our approach requires computing expectations over the prior and we compute them by sampling from that prior only once for each query $Q_k$.

\noindent\textbf{Example $f$s.} We present below three useful examples of alignment metrics $f$ that we use in our experiments. Roughly, they correspond to inducing the same answers to trajectory comparisons, mapping to a (canonically) shaped version of the same reward function (EPIC distance), as well as inducing the same ranking over trajectories.

First, we consider an $f$ based on loglikelihood, a popularly used metric in preference-based reward learning \cite{biyik2020active,wilde2021learning,biyik2023active}. Under this metric, two reward functions align with each other if human response predictions under one of them get high probabilities under the other:
\begin{align}
    f^{\textrm{LL}}(R_\weights,\! R_{\weights'}) &\!=\! g^{\textrm{LL}}(R_\weights,\! R_{\weights'}) \!+\! g^{\textrm{LL}}(R_{\weights'},\! R_\weights)\\
    g^{\textrm{LL}}(R_\weights,\! R_{\weights'}) &\!=\! \sum_{Q \in \mathcal{Q}} \!\log\! P(q\!=\!\argmax_{\xi\in Q} R_{\weights'}(\xi) \!\mid\! Q, R_{\weights})\nonumber
\end{align}
for some set of queries $\mathcal{Q}$. Ideally, this set should be representative of the environment the learned reward will be deployed to. For example, if the human will give their preferences on a simulator but the learned reward will be deployed on a real robot, the search space of the querying optimization is the simulator trajectories, but $\mathcal{Q}$ should consist of real robot trajectories. We call this variant of our querying policy $\pi^{\textrm{LL}}$.

Secondly, distance functions developed to measure the misalignment between reward functions are a natural fit for $f$. We use EPIC distance \cite{gleave2021quantifying} to measure alignment in $\pi^{\textrm{EPIC}}$:
\begin{align}
    f^{\textrm{EPIC}}(R_\weights, R_{\weights'}) = -\textrm{EPIC}(R_\weights, R_{\weights'})\:.
\end{align}
We refer to \cite{gleave2021quantifying} for the details on computing EPIC distance.

Similarly, \citet{balakrishnan2020efficient} developed a technique called $\rho$-projection to project reward functions to a space in which L2-distance can be used as a misalignment measure between those functions:
\begin{align}
    &f^{\rho}(R_\weights,R_{\weights'}) = -\norm{\rho(R_\weights), \rho(R_{\weights'})}_2\\
    &\rho(R_\omega) = \frac{[\exp R_\weights(\xi_1),\exp R_\weights(\xi_2),\dots,\exp R_\weights(\xi_N)]}{\sum_{i=1}^N \exp R_\weights(\xi_i)}\nonumber
\end{align}
for some trajectories $\xi_1,\xi_2,\dots,\xi_N$. Intuitively, $\rho$-projection metric compares how two reward functions rank the given set of trajectories under the Boltzmann rational model. Similar to the query set $\mathcal{Q}$ in $\pi^{\textrm{LL}}$, this trajectory set should be representative of the deployment environment. We denote the querying policy that uses this alignment metric with $\pi^{\rho}$.

Having presented our active querying policy and example alignment metrics, we will analyze some of its useful properties in the next section.

\section{Analysis}
We will make two remarks in our analysis. First, our method can be seen as a generalization of the mutual information based method. Second, for choices of $f$ that makes $\mathbb{E}_{\weights,\weights'\sim P(\weights\mid \mathcal{D}_k)}\left[f(R_\weights,R_{\weights'})\right]$ adaptive monotone and adaptive submodular over the sets $\mathcal{D}_k$, our method is a near-optimal solution to the problem stated in \eqref{eq:proxy_objective}.

\begin{remark}
    Slightly abusing the notation to let $f$ depend on $\mathcal{D}_k$, the mutual information based policy $\pi^\textrm{MI}$ can be seen as a special case of our approach, since it is the solution to \eqref{eq:final_objective} when $f(R_\weights,R)=\log P(\weights\mid\mathcal{D}_k)$.
    
    \begin{proof}
        Plugging $\log P(\weights\mid\mathcal{D}_k)$ in \eqref{eq:final_objective}, we get
        \begin{align*}
            \argmax_Q \!\sum_{q\in Q}\! \frac{\mathbb{E}_{\weights,\weights'}\!\left[P(q \!\mid\! Q,\! R_{\weights})P(q \!\mid\! Q,\! R_{\weights'})\log\! P(\weights\!\mid\!\mathcal{D}_k)\right]}{\mathbb{E}_{\weights''} P(q \!\mid\! Q, R_{\weights''})}
        \end{align*}
        We separate the $\weights$ and $\weights'$ terms in the numerator, and note $\weights'$ term is equivalent to the expectation in the denominator:
        \begin{align*}
            &\argmax_Q \sum_{q\in Q} \mathbb{E}_{\weights\sim P(\weights \mid \mathcal{D}_{k-1})}\left[P(q \mid Q, R_{\weights})\log P(\weights\mid\mathcal{D}_k)\right]\\
            &=\argmax_Q \mathbb{E}_{\weights,q\sim P(\weights,q \mid \mathcal{D}_{k-1},Q)}\left[\log P(\weights\mid\mathcal{D}_k)\right]\\
            &=\argmax_Q \mathbb{E}_{q\sim P(q \mid \mathcal{D}_{k-1},Q)}\left[\mathbb{E}_{\weights\sim P(\weights \mid \mathcal{D}_{k})}\left[\log P(\weights\mid\mathcal{D}_k)\right]\right]
        \end{align*}
        where we used the fact that $\mathcal{D}_k=(\mathcal{D}_{k-1},Q,q)$. Noting the inner expectation is the posterior entropy and the prior entropy $H(\weights\mid\mathcal{D}_{k-1})$ does not depend on the optimization variable $Q$, we equivalently write: 
        \begin{align*}
            &\argmax_Q \mathbb{E}_{q\sim P(q \mid \mathcal{D}_{k-1},Q)}\left[H(\weights\!\mid\!\mathcal{D}_{k-1}) - H(\weights\!\mid\!\mathcal{D}_{k-1},Q,q)\right]\\
            &= \argmax_Q I(q;\weights \mid Q, \mathcal{D}_{k-1})\:,
        \end{align*}
        which is equal to the optimization in Equation~\eqref{eq:mutual_information_objective}.
    \end{proof}
\end{remark}

Generalizing the mutual information based solution introduces some desirable theoretical properties. \citet{biyik2022learning} noted that there is no known theoretical guarantee for $\pi^{\textrm{MI}}$. However, for adaptive monotone and adaptive submodular objectives, our method enjoys the following guarantee.

\begin{remark}
    If $\mathbb{E}_{\weights,\weights'\sim P(\weights\mid \mathcal{D}_k)}\left[f(R_\weights,R_{\weights'})\right]$ is adaptive monotone and adaptive submodular over the sets $\mathcal{D}_k$, the greedy solution we presented in \eqref{eq:final_objective} will, in expectation, achieve at least $(1-\frac{1}{\epsilon})\textrm{OPT}_K$ improvement over the objective after $K$ queries, where $\textrm{OPT}_K$ is the theoretical upper bound (due to the optimal but intractable adaptive policy).
    \begin{proof}
        Directly follows from \citet{golovin2011adaptive}.
    \end{proof}
\end{remark}

While common acquisition functions in active reward learning do not satisfy adaptive submodularity, \citet{golovin2011adaptive} discuss some possibilities, including works that use adaptive submodular objectives in active learning, e.g., \cite{golovin2010near,gowtham2010modified}. In the next section, we empirically demonstrate how our method, which is tailored to the objective $f$ of the application, achieves the best results against the baselines including the mutual information optimization.

\section{Experiments}\label{sec:experiments}
We conduct experiments in three different domains: a synthetic environment, Assistive Gym that simulates an assistive robot \cite{erickson2019assistive}, and a natural language processing (NLP) task with datasets curated from Reddit \cite{ethayarajh2022understanding}. Following prior work, we use a probabilistic human model to simulate human responses to the preference queries \cite{biyik2019asking,wilde2020active,myers2021learning}.

\subsection{Human Response Model}
We simulate the human response $q$ to query $Q$ using a probabilistic model conditioned on the reward function. For this, we use the standard Boltzmann rational model, parameterized by a rationality coefficient $\beta$:
\begin{equation}
P(q = \xi \mid Q, R) = \frac{\exp{\beta \cdot R(\xi)}}{\sum_{\xi' \in Q} \exp{\beta \cdot R(\xi')}}
\end{equation}
for any trajectory $\xi\in Q$. For our experiments, we tune $\beta$ such that around 95\% of the simulated responses align with the reward functions as in \eqref{eq:preference_definition}.

\subsection{Metrics}
We claim that when the alignment function is $f$, one should use our active querying method with $\pi^f$. Therefore, we use three metrics each of which corresponds to one of the methods: loglikelihood, EPIC distance, and $\rho$-projection distance. We expect each variant of our algorithm to be the most successful under the corresponding metric.

\section{Results}
\subsection{Synthetic Environment}
We first evaluate our methods in comparison with $\pi^{\textrm{MI}}$ on a synthetic environment. This experiment demonstrates the data-efficiency of our methods in learning a reward function that can be transferred to new domains.

For this, we simulate trajectories from a source and a target domain. Trajectories in the source domain have 15 features sampled i.i.d. from $\mathcal{N}(0,1)$. To simulate the distribution shift between different domains, we let $10$ of the features of the target domain have a mixture distribution $\frac12 \mathcal{N}(-1,10^{-4})+\frac12 \mathcal{N}(1,10^{-4})$, and let the remaining $5$ features have the same distribution as in the source domain. The reward function is a linear combination of these features. The parameters to learn $\weights\in\mathbb{R}^{15}$ correspond to the weights of the features in the reward function.

\begin{figure}[t]
\includegraphics[width=0.24\textwidth]{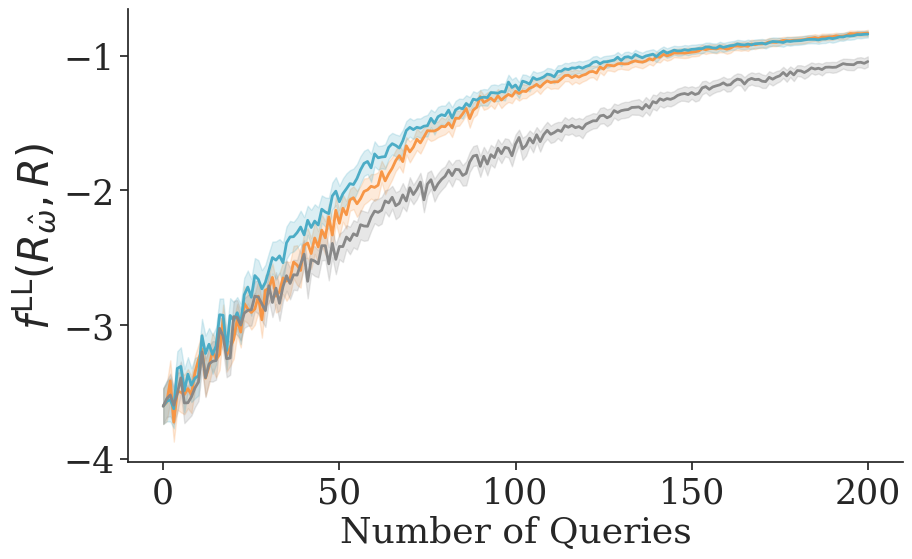}
\includegraphics[width=0.24\textwidth]{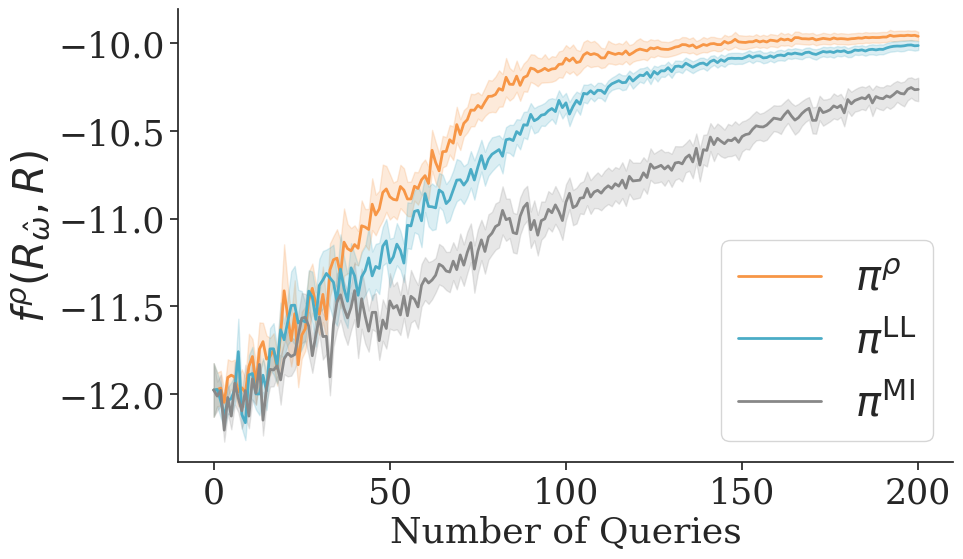}
\vspace{-15px}
\caption{Results of the synthetic environment experiment over 50 seeds (mean$\pm$se).}
\vspace{-20px}
\label{fig:test}
\centering
\end{figure}

We randomly generate $50$ different true reward parameters $w^*$ to evaluate the methods. This enables us to compute metrics against the true reward function $R_{\weights^*}$. This procedure, as well as the linear reward structure, is common in preference-based reward learning literature \cite{sadigh2017active,biyik2019asking,wilde2020active}. We compare $\pi^{\textrm{MI}}$, $\pi^{\textrm{LL}}$, $\pi^{\rho}$, and exclude $\pi^{\textrm{EPIC}}$ as it depends on more granular information in the trajectories, e.g., state-action-next state tuples, which do not exist in the synthetic data.

Figure~\ref{fig:test} shows the results of this experiment. Both $\pi^\rho$ and $\pi^{\textrm{LL}}$ significantly outperform $\pi^{\textrm{MI}}$ in both loglikelihood and $\rho$-projection based alignment metrics. These results strongly support the hypothesis that $\pi^{\textrm{MI}}$ is suboptimal when the learned reward is deployed in a different environment than the training (source) environment. It also supports the argument that we should use the variant of our algorithm that corresponds to the metric we want to optimize: $\pi^\rho$ outperforms all other methods on the $\rho$-projection based metric, and $\pi^{\textrm{LL}}$ outperforms all others on the loglikelihood.

\begin{figure*}[t]
\includegraphics[width=0.32\textwidth]{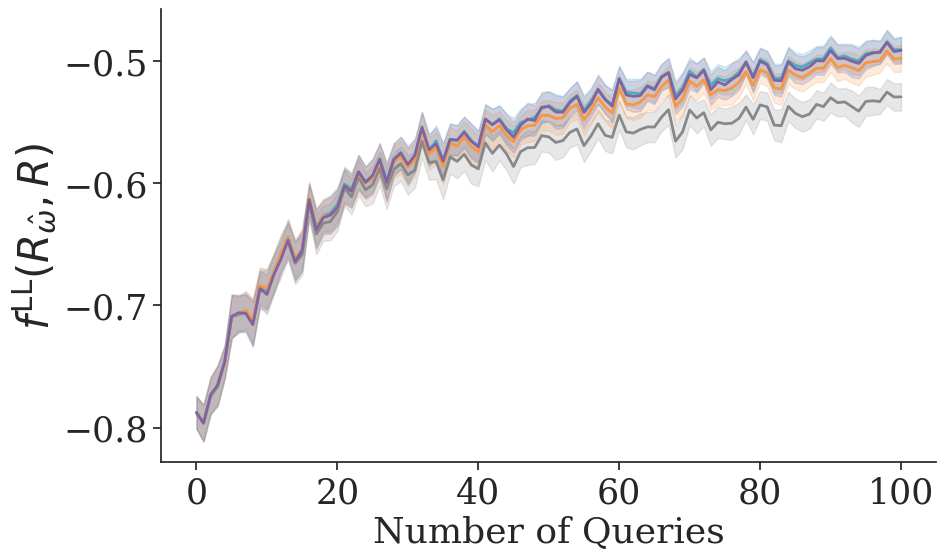}
\includegraphics[width=0.32\textwidth]{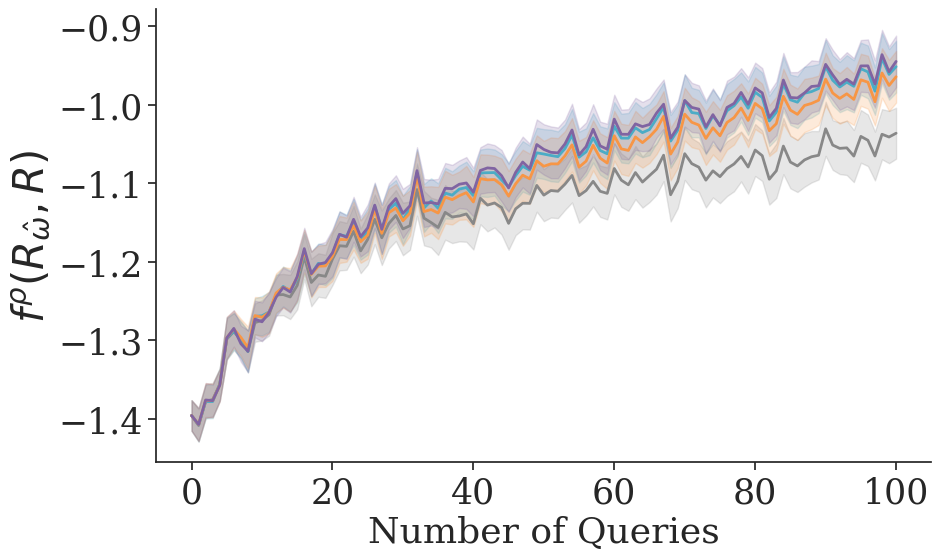}
\includegraphics[width=0.32\textwidth]{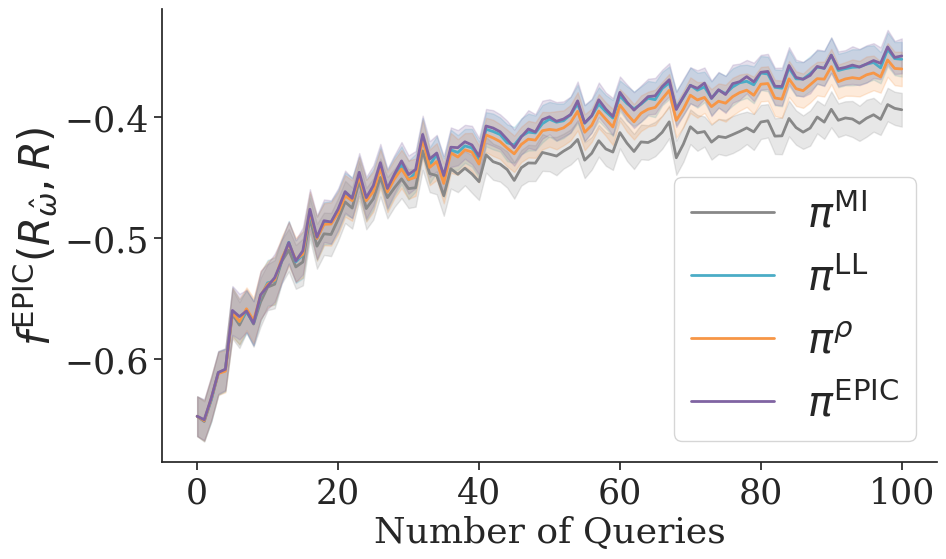}
\vspace{-5px}
\caption{Results of the Assistive Gym experiment over 100 seeds (mean$\pm$se).}
\vspace{-20px}
\label{fig:agym}
\centering
\end{figure*}

\subsection{Assistive Gym}
Next, we evaluate our methods in the Assistive Gym \cite{erickson2019assistive} simulated robotics environment. This experiment demonstrates the ability of our methods to learn a nonlinear reward that can be transferred between realistic robotics domains. We consider a robotic arm feeding a patient. The robotic arm must learn where to place the spoon, which is attached to the end effector, by asking preference queries. The experimenters have access to a Sawyer robotic arm (Rethink Robotics) but wish to learn a reward that applies to a Jaco arm (Kinova). Therefore, in this setting, the source domain involves a Sawyer arm and the target domain involves a Jaco arm. 

We randomly sample $20$ different goal positions for the spoon and let the reward function be the negative distance between the end-effector and the goal positions. However, the goal is not known by the robot and must be learned via preference queries. Hence, the learnable parameters of the reward function $\weights$ correspond to the goal position.

Figure~\ref{fig:agym} shows the results of this experiment. $\pi^{\textrm{LL}}$, $\pi^{\textrm{EPIC}}$, and $\pi^{\rho}$ all outperform $\pi^{\textrm{MI}}$ in loglikelihood, rho-projection alignment, and EPIC-distance alignment score, showing that our algorithm succeeds in learning nonlinear rewards for the target domain. The inefficiency of $\pi^{\textrm{MI}}$ at learning this simple nonlinear reward suggests that taking the deployment environment into account is essential to solving more complex active learning problems.

\subsection{NLP Task}
Finally, we evaluate our methods in an NLP task. This is inspired by InstructGPT \cite{ouyang2022training}, which fine-tunes a large language model with preferences, a popular practice in NLP.

We consider a setting where a human user may easily compare the quality of texts in one domain, but it is costly in some other domains. For example, texts in the target domain may be much longer than those in the source domain, or the user may be less knowledgeable about the target domain. Both of these cases have been pointed out as limitations of reinforcement learning from human feedback \cite{casper2023open}, but as we will show, our methods alleviate these problems by enabling data-efficient reward learning on a simpler source domain.

For this, we employ Stanford Human Preferences Dataset \cite{ethayarajh2022understanding}, which has curated data from Reddit. Specifically, we take r/askvet and r/askphilosophy subreddits, which contain discussions on completely different topics. Our goal is to learn a reward function for writing quality by using preferences on r/askvet and then check if the learned reward aligns with the true preferences in r/askphilosophy. To this end, we model each comment in the subreddits as a trajectory $\xi$, and we label each of them with: sentiment analysis over ``emotion'', ``hate'', ``irony'', ``offensive'', ``sentiment'' \cite{barbieri2020tweeteval}, Flesch-Kincaid grade level \cite{flesch1949art}, Flesch-Kincaid reading ease \cite{flesch1949art}, Dale-Chall readability \cite{dale1948formula}, Coleman–Liau index \cite{coleman1975computer}, automated readability index \cite{senter1967automated}, and the relevance between the comment and its main post according to the model by \citet{liu2021generalizing}. The reward of each comment is then a linear combination of these features after normalization.

We randomly generate 50 parameter vectors $w^*$, representing different views on writing quality. We restrict the query space such that we can query the user only with comments that belong to the same main post. 

\begin{figure}[t]
\includegraphics[width=0.24\textwidth]{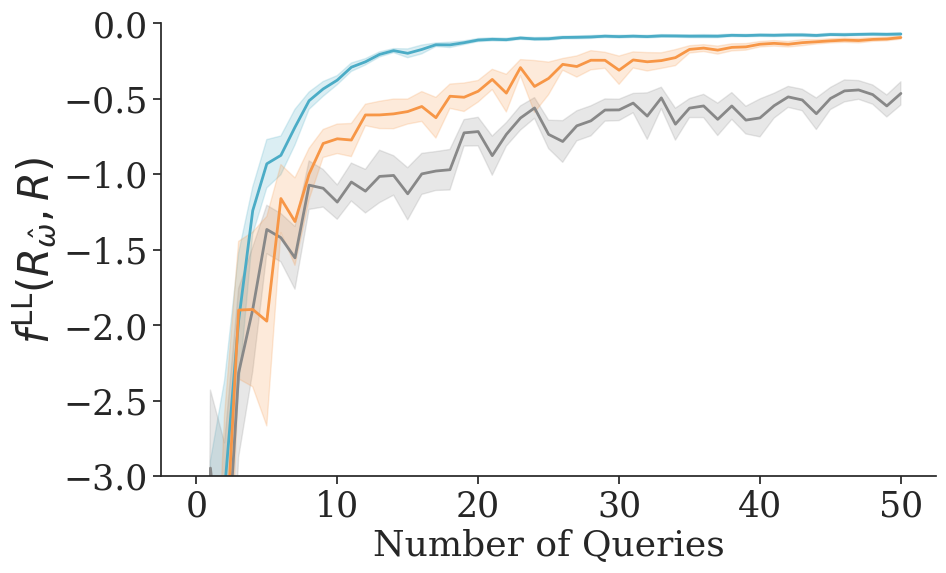}
\includegraphics[width=0.24\textwidth]{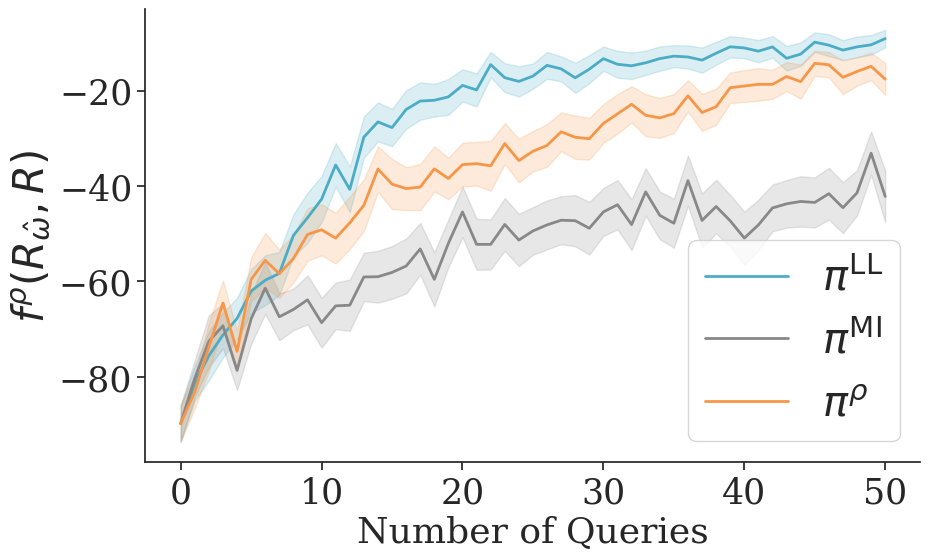}
\vspace{-15px}
\caption{NLP experiments results over 50 seeds (mean$\pm$se).}
\vspace{-20px}
\label{fig:reddit}
\centering
\end{figure}

Results from this experiment are shown in Figure~\ref{fig:reddit}. Both $\pi^\rho$ and $\pi^{\textrm{LL}}$ outperform $\pi^{\textrm{MI}}$ in the log-likelihood and $\rho$-projection based alignment metrics ($\pi^{\textrm{EPIC}}$ was excluded as there is no granular information about the trajectories). Surprisingly, $\pi^{\textrm{LL}}$ performs better than $\pi^\rho$ even when the alignment metric is $f^\rho$. Noting that this is not the case early in the training with a smaller number of queries, we posit this may be because of the greedy approximation to the original optimization problem (see Equation~\eqref{eq:greedy_objective}). It is also possible that certain alignment metrics are better suited for some domains than others.

\section{Conclusion}
We introduced a new method for active preference-based learning of a reward that behaves similarly to the true reward in terms of a user-defined alignment metric. We have shown results comparing our method using three different alignment metrics with the state-of-the-art baseline on various environments. The results demonstrated the advantages of our method in learning both linear and nonlinear rewards.

Future works may investigate different alignment metrics, and their implications on the learned rewards. They may also explore how our methodology can be extended to gradient-based learning methods (as opposed to Bayesian) so that it can be applied to settings where rewards are modeled with a large number of parameters, e.g., deep neural networks.

\section*{Acknowledgments}
This work was supported by Cocosys SRC center, and an ONR YIP. The authors also acknowledge a gift from Open Philanthropy to support the work of the Center for Human-Compatible AI at UC Berkeley, 

\section*{APPENDIX} 
\subsection{Derivation of Equation~\eqref{eq:final_objective}}
We start from Equation~\eqref{eq:greedy_objective} and expand the expectations, where integrals are over the entire parameter space for $\weights$:
\begin{align*}
    \argmax_Q \sum_{q\in Q}\int\int &P(q \mid Q, \mathcal{D}_{k-1})P(\weights \mid \mathcal{D}_{k-1}, Q, q)\\
    &P(\weights' \mid \mathcal{D}_{k-1}, Q, q)f(R_\weights, R_{\weights'})d\weights d\weights'\:.
\end{align*}
Using Bayes rule, $\weights \perp Q \mid \mathcal{D}_{k-1}$, and $q \perp \mathcal{D}_{k-1} \mid \weights, Q$ to replace $P(\weights \mid \mathcal{D}_{k-1}, Q, q)$ with $\frac{P(\weights \mid \mathcal{D}_{k-1})P(q \mid \weights, Q)}{P(q \mid \mathcal{D}_{k-1}, Q)}$, we get
\begin{align*}
    \argmax_Q \sum_{q\in Q} & \frac{1}{P(q \mid \mathcal{D}_{k-1}, Q)}\!\int\!\int\! P(\weights \!\mid\! \mathcal{D}_{k-1})P(\weights' \!\mid\! \mathcal{D}_{k-1})\\
    &P(q \mid \weights, Q)P(q \mid \weights', Q) f(R_\weights, R_{\weights'})d\weights d\weights'\:.
\end{align*}
Rewriting the integrals as expectations gives
\begin{align*}
    \argmax_Q \sum_{q\in Q} & \frac{\mathbb{E}_{\weights,\weights'\mid\mathcal{D}_{k-1}}\!\left[P(q \!\mid\! \weights, Q)P(q \!\mid\! \weights', Q)f(R_\weights, R_{\weights'})\right]}{P(q \mid \mathcal{D}_{k-1}, Q)}
\end{align*}
Finally, we note $P(q \mid \mathcal{D}_{k-1}, Q) \!=\! \int P(\weights, q \!\mid\! \mathcal{D}_{k-1}, Q)d\weights \!=\! \int P(\weights \!\mid\! \mathcal{D}_{k-1})P(q \!\mid\! Q, R_{\weights})d\weights$. Plugging this final expression as an expectation into the objective, we reach the final objective we presented in \eqref{eq:final_objective}:
\begin{align*}
    \argmax_Q \sum_{q\in Q} & \frac{\mathbb{E}_{\weights,\weights'\mid\mathcal{D}_{k-1}}\!\left[P(q \!\mid\! \weights, Q)P(q \!\mid\! \weights', Q)f(R_\weights, R_{\weights'})\right]}{\mathbb{E}_{\weights''\mid \mathcal{D}_{k-1}}P(q \mid Q, R_{\weights''})}
\end{align*}

\balance
\renewcommand*{\bibfont}{\small}
\printbibliography

\end{document}